\documentclass[10pt,twocolumn,letterpaper]{article}

\usepackage{iccv}
\usepackage{times}
\usepackage{epsfig}
\usepackage{graphicx}
\usepackage{amsmath}
\usepackage{amssymb}
\usepackage{upgreek}
\usepackage{bm}
\usepackage{booktabs}

\usepackage[pagebackref=true,breaklinks=true,letterpaper=true,colorlinks,bookmarks=false]{hyperref}

\iccvfinalcopy 

\def\iccvPaperID{3972} 

\ificcvfinal\fi

\begin{document}

\title{GistNet: a Geometric Structure Transfer Network for Long-Tailed Recognition}

\author{Bo Liu\\
UC, San Diego\\
{\tt\small boliu@ucsd.edu}
\and
Haoxiang Li\\
Wormpex AI Research\\
{\tt\small lhxustcer@gmail.com}
\and
Hao Kang\\
Wormpex AI Research\\
{\tt\small haokheseri@gmail.com}
\and
Gang Hua\\
Wormpex AI Research\\
{\tt\small ganghua@gmail.com}
\and
Nuno Vasconcelos\\
UC, San Diego\\
{\tt\small nuno@ece.ucsd.edu}
}

\maketitle
\ificcvfinal\thispagestyle{empty}\fi

\begin{abstract}
The problem of long-tailed recognition, where the number of examples per class is highly unbalanced, is considered. It is hypothesized that the well known tendency of standard classifier training to overfit to popular classes can be exploited for effective transfer learning. Rather than eliminating this overfitting, e.g. by adopting popular class-balanced sampling methods, the learning algorithm should instead leverage this overfitting to {\it transfer\/} geometric information from popular to low-shot classes. A new classifier architecture, GistNet, is proposed to support this goal, using constellations of classifier parameters to encode the class geometry. A new learning algorithm is then proposed for {\it GeometrIc Structure Transfer\/} (GIST),  with resort to a combination of loss functions that combine class-balanced and random sampling to guarantee that, while overfitting to the popular classes is restricted to geometric parameters, it is leveraged to transfer class geometry from popular to few-shot classes. This enables better generalization for few-shot classes without the need for the manual specification of class weights, or even the explicit grouping of classes into different types. 
Experiments on two popular long-tailed recognition datasets show that GistNet outperforms existing solutions to this problem.
\end{abstract}

\section{Introduction}
The availability of large-scale datasets, with large numbers of images per class~\cite{imagenet_cvpr09}, has been a major factor in the success of deep learning for tasks such as object recognition. However, these datasets are manually curated and artificially balanced. This is unlike most real world applications, where the frequencies of examples from different classes can be highly unbalanced, leading to skewed distributions with long tails. 

This has motivated recent interest in the problem of long-tailed recognition~\cite{liu2019large}, 
where the training data is highly unbalanced but the test set is kept balanced, so that equally good performance on all classes is crucial to achieve high overall accuracy. 
\begin{figure*}[t!]
\centering{
	\includegraphics[width=0.95\textwidth]{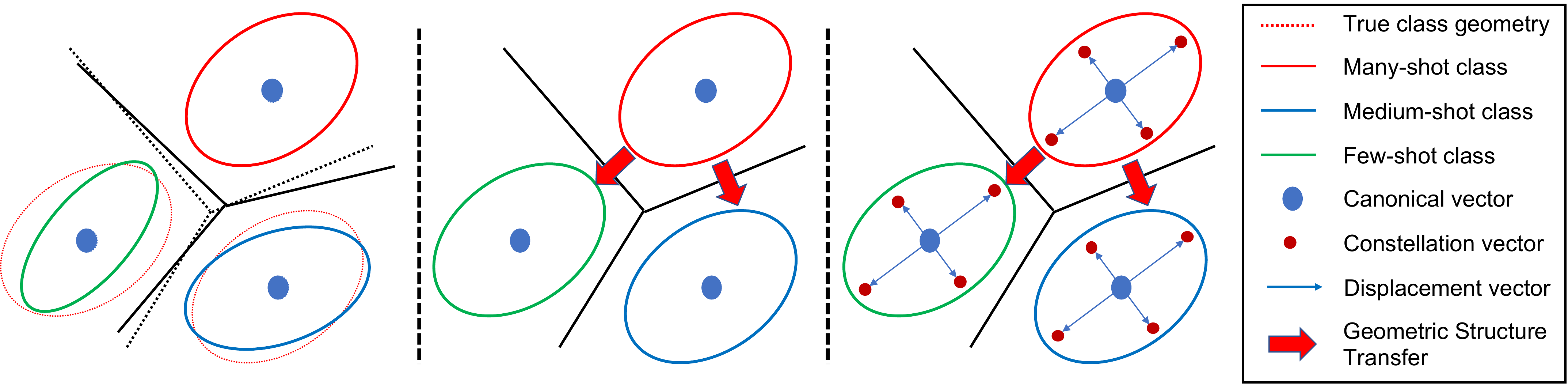}
	}
	\caption{Left: in long-tailed recognition, the small number of samples from medium- and few-shot classes make it difficult to learn their geometry, leading to inaccurate class boundaries. This is unlike many-shot classes, whose natural geometry can usually be learned. Middle: the boundaries are corrected by transferring the geometric structure of the many-shot classes to the classes with few examples. Right: GistNet implements geometric structure transfer by implementing constellations of classification parameters. These consist of a class-specific center and a set of displacements shared by all classes. Under GIST training, these tend to follow the natural geometry of the many-shot classes, which is transferred to the medium- and few-shot classes.}
	\label{fig:motivation}
\end{figure*}

Success in the long-tailed recognition setting requires specific handling of class imbalance during training, since a classifier trained with the standard cross-entropy loss will overfit to highly populated classes and perform poorly on low-shot classes. 
This has motivated several works to fight class overfitting with  methods, like data re-sampling~\cite{zhang2017range} or cost-sensitive losses~\cite{lin2017focal}, that place more training emphasis on the examples of lower populated classes. 

It is, however, difficult to design augmentation or class weighting schemes that do not either under or over-emphasize the few-shot classes. In this work, we seek an approach that is fully data driven and {\it leverages\/} overfitting to the popular classes, rather than combat it. The idea is to transfer some properties of these classes, which are well learned by the standard classifier, to the classes with insufficient data, where this is not possible.

For this, we leverage the interpretation of a deep classifier as the composition of an embedding, or feature extractor, implemented with several neural network layers and a parametric classifier, implemented with a logistic regression layer, at the top of the network.  While the embedding is shared by all classes, the classifier parameters are class-specific, namely a weight-vector per class, as shown in Figure~\ref{fig:motivation}. 

We exploit the fact that the configuration of these weight vectors determines the geometry of the embedding. This consists of the class-conditional distribution, and associated metric, of the feature vectors of each class, which define the class boundaries. For a well learned network, this geometry is identical for all classes.  In the long-tailed setting, the geometry is usually well learned for many-shot classes, but not for classes with insufficient training samples, as shown in the left of Figure~\ref{fig:motivation}. 

The goal is to transfer the geometric structure of the many-shot classes to the classes with few examples, as shown in the middle of the figure, to eliminate this problem. The challenge is to implement this transfer using only the available training data, i.e. without manual specification of class-weights or heuristic recipes, such as equating these weights to class frequency.

We address this challenge with a combination of contributions. First, we enforce a globally learned geometric structure, which is shared by all classes. To avoid the complexity of learning a full-blown distance function, which frequently requires a large covariance matrix, we propose a structure composed by a constellation of classifier parameters, as shown on the right of Figure~\ref{fig:motivation}. This consists of a class-specific center, which encodes the location of the class, and a set of displacements, which are shared by all classes and encode the class geometry. 

Second, we rely on a mix of randomly sampled and class-balanced mini-batches to define two losses that are used to learn the different classifier parameters. Class-balanced sampling is
used to learn the class-specific center parameters. This guarantees that the learning is based on the same number of examples for all classes, avoiding a bias towards larger classes. Random sampling is used to learn the shared geometry parameters (displacements). This leverages the tendency of the standard classifier to overfit to the popular classes, making them dominant for the learning of class geometry, and thus allowing the transfer of geometric structure from these to the few-shot classes. In result, the few shot classes are learned equally to the popular classes with respect to location but inherit their geometric structure, which enables better generalization. 

We propose a new learning algorithm, denoted {\it GeometrIc Structure Transfer\/} (GIST), that combines the two types of sampling, so as to naturally account for all the data in the training set, without the need for the manual specification of class weights, or even the explicit grouping of classes into different types. While we adopt the standard division into many-, medium-, and few-shot classes for evaluation, this is not necessary for training. 

A deep network that implements the parameter constellations of Figure~\ref{fig:motivation} and GIST training is then introduced and denoted as the GistNet. Experiments on two popular long-tailed recognition datasets show that it outperforms previous approaches to long-tailed recognition.

Overall, this work makes several contributions to long-tailed recognition. First, we point out that the tendency of the standard classifier to overfit to popular classes can be advantageous for transfer learning. The goal should not be to eliminate this overfitting, e.g. by uniquely adopting the now popular class-balanced sampling, but leverage it to transfer geometric information from the popular to the low-shot classes. 

Second, we propose a new GistNet classifier architecture to support this goal, using constellations of classifier parameters to encode the class geometry. 

Third, we introduce a new learning algorithm, GIST, that combines class-balanced and random sampling to leverage overfitting to the popular classes and enable the transfer of class geometry from popular to few-shot classes.

\begin{figure}[t!]
\centering{
	\includegraphics[width=0.45\textwidth]{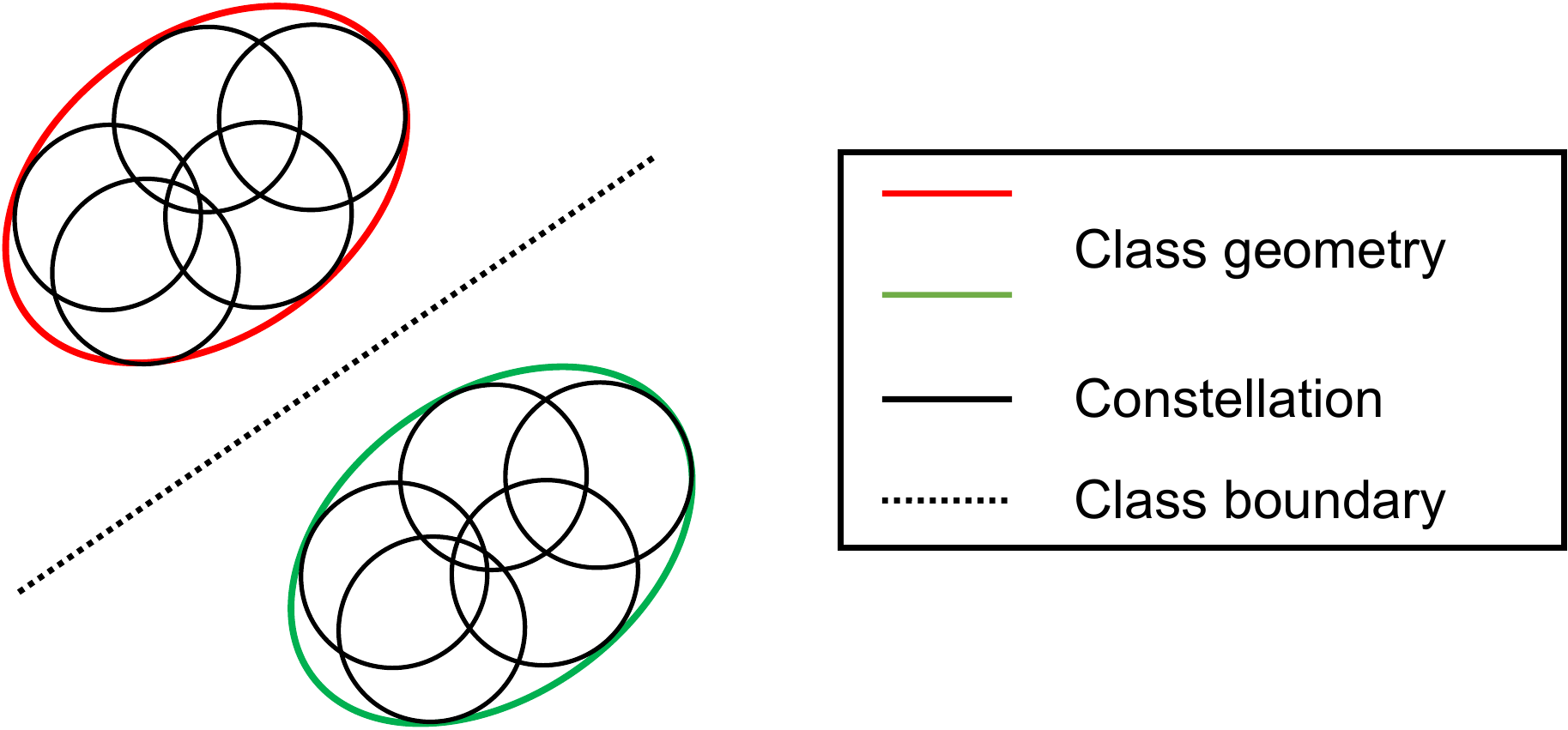}
	}
	\caption{GistNet approximates the shared geometry by a constellation (mixture) of spherical Gaussians.}
	\label{fig:geometry}
\end{figure}

\begin{figure}[t!]
\centering{
	\includegraphics[width=0.45\textwidth]{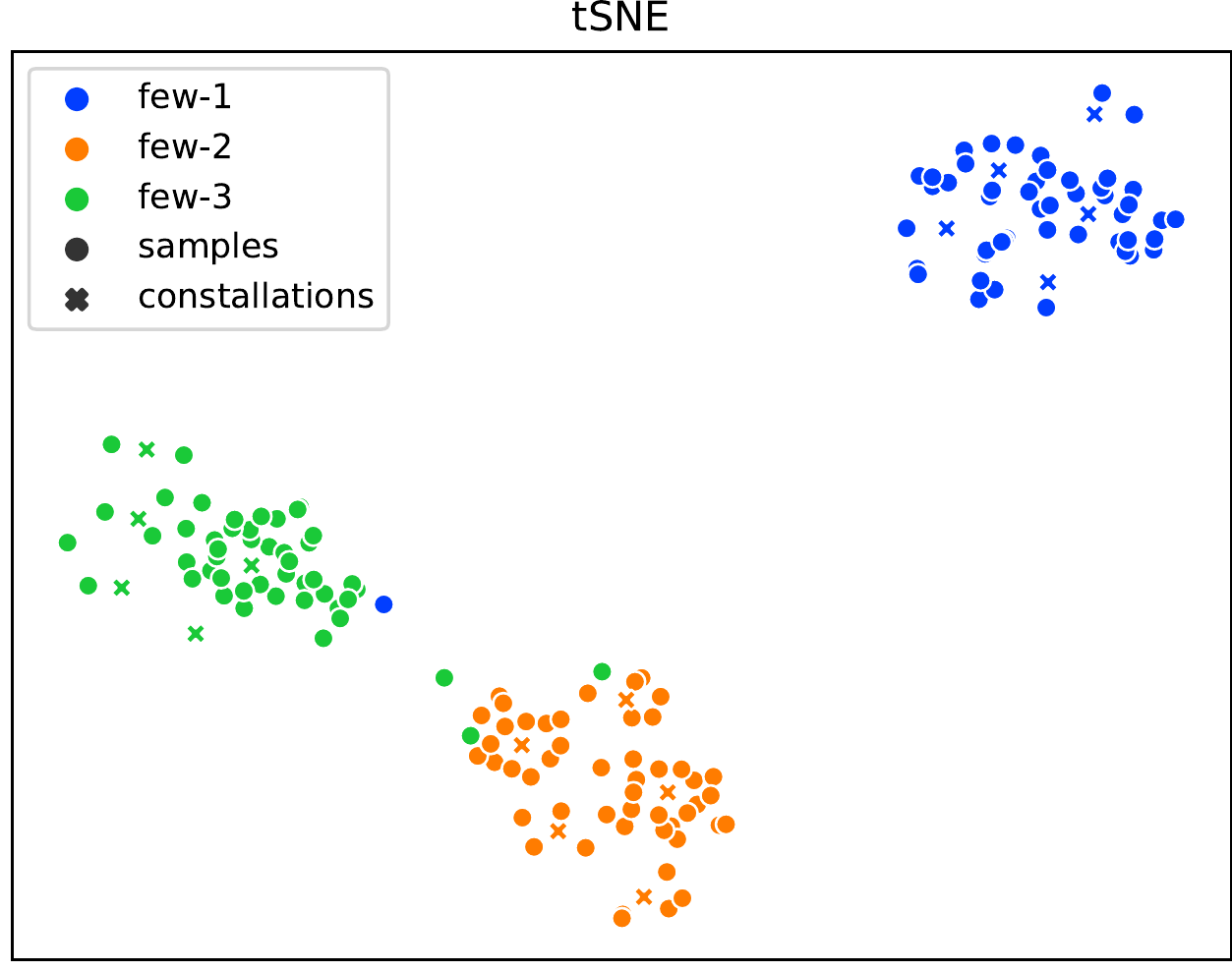}
	}
	\caption{t-SNE visualization of 3 few-shot classes on ImageNet-LT test set, together with the constellations $\mathbf{w}_{kj}$.}
	\label{fig:constellation}
\end{figure}

\section{Related Work}
\noindent
{\bf Long-tailed recognition}
has received increased attention in the recent past~\cite{wang2017learning,oh2016deep,lin2017focal,zhang2017range,liu2019large,wang2016learning}. Several approaches have been proposed, including metric learning~\cite{oh2016deep,zhang2017range}, hard negative mining~\cite{lin2017focal}, or meta-learning~\cite{wang2016learning}. Some of these rely on novel loss functions, such as the lift loss~\cite{oh2016deep}, which introduces margins between many training samples, the range loss~\cite{zhang2017range}, which encourages data in the same class (different classes) to be close (far away), or the focal loss~\cite{lin2017focal}, which conducts online hard negative mining. These methods tend to improve performance on the few-shot end at the cost of many-shot accuracy. 

Other methods, e.g. class-balanced experts~\cite{sharma2020long} and knowledge distill~\cite{xiang2020learning}, try to avoid this problem by manually dividing the training data into subsets, based on the number of examples, and training an expert per subset. However, experts learned from arbitrary data divisions can be sub-optimal, especially for few-shot classes. 

Kang et al.~\cite{kang2019decoupling} tackles the data-imbalance problem by decoupling the training feature embedding and classifier. Zhou et al.~\cite{zhou2020bbn} also shows the effectiveness by using different training strategies on feature embedding and classifier, and achieves this by cumulative learning. These methods, however, do not discuss the class geometry problem. In face recognition, Liu et al.~\cite{liu2020deep} explores the long-tailed problem by knowledge transfer. The idea is similar to ours. But they achieve this by data synthesis, while we rely on model design and training strategy. 

GistNet is closest to the OLTR approach of~\cite{liu2019large}, which uses a visual memory and attention to propagate information between classes. This, however, is insufficient to guarantee the transfer of geometric class structure, as intended by GIST.

\noindent
{\bf Few-shot learning} is a well-researched problem. A popular group of approaches is based on meta-learning, using gradient based methods such as MAML and its variants~\cite{finn2017model,finn2018probabilistic}, or LEO~\cite{rusu2018meta}. These methods take advantage of second derivatives to update the model from few-shot samples. Alternatively, the problem has been addressed with metric based solutions, such as the matching~\cite{vinyals2016matching}, prototypical~\cite{snell2017prototypical}, and relation~\cite{sung2018learning} networks. These approaches learn metric embeddings that are transferable across classes. 

There have also been proposals for feature augmentation,  aimed to augment the data available for training, e.g. by combining GANs with meta-learning~\cite{wang2018low}, synthesizing features across object views~\cite{Liu_2018_CVPR} or other forms of data hallucination~\cite{hariharan2017low}. All these methods are designed specifically for few-shot classes and often under-perform for many-shot classes.

\noindent
{\bf Learning without forgetting} aims to train a model sequentially on new tasks without forgetting the ones already learned. This problem has been recently considered in the few-shot setting~\cite{gidaris2018dynamic}, where the sequence of tasks includes a mix of many-shot and few-shot classes. 

Proposed solutions~\cite{gidaris2018dynamic,ren2019incremental} attempt to deal with this 
by training on many-shots first, using the many-shot class weights to generate few-shot class weights, and combining them together. These techniques are difficult to generalize to long-tailed recognition, where
the transition from many- to few- shot classes is continuous and includes a large number of medium-shot classes.

\section{Geometric Structure Transfer}

In this section, we introduce the proposed solution of the long-tailed recognition problem by geometric structure transfer and the GistNet architecture.

\begin{figure*}[t!]
\centering{
	\includegraphics[width=0.95\textwidth]{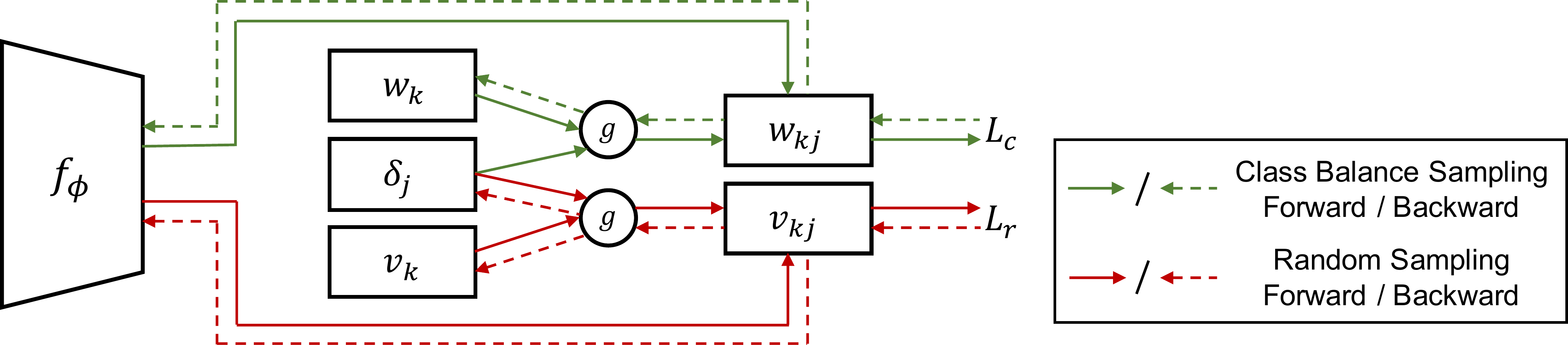}
	}
	\caption{GIST training. Solid arrows represent feed-forward and dashed ones back-propagation. Class-balanced mini-batches are used for the green connections, to guarantee that the parameters $\mathbf{w}_k$ are class-specific. Random sampling mini-batches are used for the red connections, enabling the displacements $\mathbf{\delta}_j$ to be learned predominantly from many-shot classes. Note that the shape parameters $\mathbf{\delta}_j$ receive no gradient from the class-balanced loss ${\cal L}_c$ and the constellation centers $\mathbf{w}_k$ receive no gradient from the random sampling loss ${\cal L}_r$.}
	\label{fig:structure}
\end{figure*}

\subsection{Regularization by Geometric Structure Transfer}
\label{sec:geometric}

A popular architecture for classification is the softmax classifier. This consists of an embedding that maps images $\mathbf{x} \in {\cal X}$ into feature vectors $f_\phi(\mathbf{x}) \in {\cal F}$, implemented by multiple neural network layers, and a softmax layer that estimates class posterior probabilities according to 
\begin{equation}
    p(y = k | \mathbf{x}; \phi, \mathbf{w}_k) = \frac{\exp[\mathbf{w}_k^Tf_\phi(\mathbf{x})]}
    {\sum_{k'}\exp[\mathbf{w}_{k'}^Tf_\phi(\mathbf{x})]}
    \label{eq:softmax}
\end{equation}
where $\phi$ denotes the embedding parameters and $\mathbf{w}_k$ is the weight vector of the $k^{th}$ class.

The model is learned with a training set $\mathbb{S} = \{(\mathbf{x}_i, y_i)\}_{i=1}^{n^s}$ of $n^s$ examples, by minimizing the cross entropy loss
\begin{equation}
    {\cal L}_{CE} = \sum_{(\mathbf{x}_i, y_i) \in \mathbb{S}} -\log p(y_i | \mathbf{x}_i).
    \label{eq:softmaxce}
\end{equation}
Recognition performance is evaluated on a test set $\mathbb{T}=\{(x_i, y_i)\}_{i=1}^{n^t}$, of $n^t$ examples.

Learning with (\ref{eq:softmaxce}) produces a particular data-driven embedding geometry, which we denote the {\it natural\/} geometry for the training data. While parameters $\mathbf{w}_k$ of the classifier is class-specific and describes class centers, it is usually impossible to determine this geometry from the learned network parameters.\footnote{See supplementary material for detail.}

This is not a problem in regular large-scale recognition. In such a case, each class has enough training data and the natural geometry is successfully learned under cross-entropy loss without further regulations. For long-tailed recognition problems the situation is different. As in few-shot learning,
the limited training data of few-shot classes leads to weakly defined class-conditional distributions and embedding geometry. However, this is not the case for classes with many samples, whose natural geometry can be learned from the data. In result, as illustrated in the left of Figure~\ref{fig:motivation}, the true class boundaries are usually not well learned for the few-shot classes. 

In this work, we seek to leverage geometric regularization to improve the learning of the few-shot classes without sacrificing performance for the populated classes. 

One possibility would be to enforce a pre-defined geometry for all classes, e.g.
by adopting Mahalanobis distance $d(f_\phi(\mathbf{x}), \mathbf{\mu}) = (f_\phi(\mathbf{x})- \mathbf{\mu})^T \Sigma^{-1}  (f_\phi(\mathbf{x})- \mathbf{\mu})$ associated with Gaussian class conditionals of covariance $\Sigma$, or by assuming Gaussian class-conditionals and regularizing the covariance to be close to a pre-defined $\Sigma$. 

This has several problems. First, it is not clear what the covariance $\Sigma$ should be. Second, it ignores the natural geometry of the popular classes, which is well learned by the classifier of~(\ref{eq:softmax}). Third,
given the large dimensionality of $f_\phi(\mathbf{x})$, covariance regularization is difficult to implement, even for classes with many examples.

To avoid these problems, we seek a learning-based solution that does not require covariance estimation
and leverages the natural geometry of the  popular classes to regularize the geometry of the few-shot classes.  
Rather than forcing geometry through a distance function,
which is hard to learn and implement, we pursue an alternative
approach to guarantee that all classes have a {\it shared} geometric structure. 

Ideally, this structure should be learned from
data, so as to 1) follow the natural geometry of the highly populated classes, and 2) allow the transfer of that geometry to the classes of few examples. It should also be encoded
in a relatively small number of parameters, which at most grows linearly with the dimension of $f_\phi(\mathbf{x})$.

To achieve these goals, we continue to rely on the softmax
classifier of~(\ref{eq:softmax}) and the cross-entropy loss of~(\ref{eq:softmaxce}), but use an alternative 
implementation of the softmax layer
\begin{equation}
\begin{aligned}
    p_\phi (y = k | \mathbf{x}) &= \frac{\exp[\max_j \mathbf{w}_{kj}^Tf_\phi(\mathbf{x})]}{\sum_{k'}\exp[\max_j \mathbf{w}_{k'j}^Tf_\phi(\mathbf{x})]}, \\
    \mathbf{w}_{kj} &= g(\mathbf{w}_k, \mathbf{\delta}_j),
    \label{eq:amplifier}
\end{aligned}
\end{equation}
where the canonical 
parameter vector $\mathbf{w}_k$ is replaced by a {\it constellation\/} of parameter vectors $\mathbf{w}_{kj}$,
which are a function of $\mathbf{w}_k$ and a set of {\it structure parameters\/} $\delta_j$ shared by all classes. Under the simplest implementation of this idea, $g(\mathbf{w}_k, \mathbf{\delta}_j) = \mathbf{w}_k + \mathbf{\delta}_j$ and the structure parameters are a 
 set of displacement vectors, as shown in the right of Figure~\ref{fig:motivation}. 
 
Since these displacements are shared by all classes, the constellation is simply replicated around each $\mathbf{w}_k$, which is learned per class.
Because, under the loss of~(\ref{eq:softmaxce}), the highly populated classes tend to dominate the optimization of the shared parameters, the displacements $\delta_j$ tend to follow the natural geometry of these classes, which is thus transferred to the few-shot classes. This regularizes the learning of these classes, enabling the recovery of the true classification boundaries, as shown in the right of Figure~\ref{fig:motivation}. 

The displacements $\delta_j$ are the parameters that contain geometry information. They transfer the geometry from highly populated classes to few-shot classes. With the help of geometry transfer, the model learns a better geometry for few-shot classes.

As shown in Figure~\ref{fig:geometry}, (\ref{eq:amplifier}) is equivalent to replacing the natural geometry by several spherical Gaussians of means $w_{kj}$ and choosing the one closest to the feature. This approximates the non-regulated geometry by a constellation of $5$ spherical Gaussians, one per $\mathbf{w}_{kj}$. This geometry is visualized in Figure~\ref{fig:constellation}, where features from different classes are regulated by class specific constellations respectively.
The constellation can be regarded as an umbrella. The model can learn the shape of the umbrella and where to place the umbrella for each class.

We denote the approach as {\it GeometrIc Structure Transfer\/} (GIST), to capture the fact that it transfers the essence, or gist, of the class geometry from popular to few-shot classes. 

Note that the classifier in (\ref{eq:amplifier}) is different from that in (\ref{eq:softmax}). There is an additional constraint: that the displacements $\mathbf{\delta}_j$ are constant across classes. To avoid the model learns $w_{k}$ to fit one of the constellations and ignore others. We first train the classifier from (\ref{eq:softmax}) to get a stable initialization of $w_{k}$, and then the whole classifier is trained to get the class-agnostic displacements. In such a case, the model will have to fit all available constellations to get lower loss instead of fitting one of them. Empirical examination in Section~\ref{sec:ablate} shows the actual usage of $\{\delta_j\}$ is decent, and supports this assumption.

\subsection{Normalization}

Recent works~\cite{gidaris2018dynamic,liu2019large} have shown that better few-shot or long-tailed classification accuracies are frequently obtained
by performing the classification on the unit sphere, i.e.
normalizing both embedding and classifier parameters to have
unit norm. We follow this practice and adopt the weighted cosine classifier~\cite{gidaris2018dynamic}, replacing (\ref{eq:amplifier}) with
\begin{equation}
\begin{aligned}
    p_\phi (y = k | \mathbf{x}) &= \frac{\exp[\max_j s_\tau(f_\phi(\mathbf{x}), \mathbf{w}_{kj})]}{\sum_{k'}\exp[\max_j s_\tau(f_\phi(\mathbf{x}), \mathbf{w}_{k'j}]},
    \\
     s_\tau(f_\phi(\mathbf{x}),\mathbf{w}) &= \tau\frac{\mathbf{w}^Tf_\phi(\mathbf{x})}{||\mathbf{w}||||f_\phi(\mathbf{x})||}
    \label{eq:cosine}
\end{aligned}
\end{equation}
where $\tau$ is a parameter that controls the smoothness of the posterior distribution. This architecture is denoted as GistNet. In our implementation, $\tau$ is randomly initialized and learned end-to-end.

\begin{table*}
  \caption{Results on ImageNet-LT and Places-LT. ResNet-10/152 are used for all methods. For many-shot $t >100$, for medium-shot $t \in (20,100]$, and for few-shot $t \leq 20$, where $t$ is the number of training samples.}
  \label{tab:imagenetlt}
  \centering
  \small
  \begin{tabular}{l|cccc|cccc}
    \toprule
     & \multicolumn{4}{c|}{ImageNet-LT} & \multicolumn{4}{c}{Places-LT} \\
    Method   & Overall & Many-Shot & Medium-Shot & Few-Shot & Overall & Many-Shot & Medium-Shot & Few-Shot \\
    \midrule
    Plain Model & 23.5 & 41.1 & 14.9 & 3.6 & 27.2 & {\bf 45.9} & 22.4 & 0.36 \\
    Lifted Loss~\cite{oh2016deep} & 30.8 & 35.8 & 30.4 & 17.9 & 35.2 & 41.1 & 35.4 & 24.0 \\
    Focal Loss~\cite{lin2017focal} & 30.5 & 36.4 & 29.9 & 16.0 & 34.6 & 41.1 & 34.8 & 22.4 \\
    Range Loss~\cite{zhang2017range} & 30.7 & 35.8 & 30.3 & 17.6 & 35.1 & 41.1 & 35.4 & 23.2 \\
    FSLwF~\cite{gidaris2018dynamic} & 28.4 & 40.9 & 22.1 & 15.0 & 34.9 & 43.9 & 29.9 & 29.5 \\
    OLTR~\cite{liu2019large} & 35.6 & 43.2 & 35.1 & 18.5 & 35.9 & 44.7 & 37.0 & 25.3 \\
    Decoupling~\cite{kang2019decoupling} & 41.4 & 51.8 & 38.8 & 21.5 & 37.9 & 37.8 & 40.7 & 31.8 \\
    Distill~\cite{xiang2020learning} & 38.8 & 47.0 & 37.9 & 19.2 & 36.2 & 39.3 & 39.6 & 24.2 \\
    \midrule
    GistNet & {\bf 42.2} & {\bf 52.8} & {\bf 39.8} & {\bf 21.7} & {\bf 39.6} & 42.5 & {\bf 40.8} & {\bf 32.1} \\
    \bottomrule
  \end{tabular}
\end{table*}

\begin{table}
  \caption{Results on the iNaturalist 2018. All methods are implemented with ResNet-50.}
  \label{tab:inat}
  \centering
  \begin{tabular}{lc}
    \toprule
    Method   & Accuracy \\
    \midrule
    CB-Focal~\cite{cui2019class} & 61.1 \\
    LDAM+DRW~\cite{cao2019learning} & 68.0 \\
    Decoupling~\cite{kang2019decoupling} & 69.5 \\
    \midrule
    GistNet & {\bf 70.8} \\
    \bottomrule
  \end{tabular}
\end{table}

\subsection{GIST Training}
Deep networks are trained by stochastic gradient descent (SGD). This randomly samples mini-batches of $b$ samples, and iterates across the training set. Due to the extreme class imbalance of long-tailed recognition, SGD tends to bias the model towards the classes with more samples. 

In the literature, this problem is usually addressed by class-balanced sampling~\cite{zhang2017range}. This first randomly samples $b_c$ classes with equal probability, and draws $b_n$ samples per class, producing a mini-batch of $b=b_c\times b_n$ samples. By iterating through all classes, the model is trained with an overall equal number of examples per class. For the classifier of (\ref{eq:softmax}), class-balanced sampling can significantly outperform regular sampling on few-shot classes. This also makes it a good solution for learning
the class specific parameters $\{\mathbf{w}_k\}$ of GistNet.  

However, the bias of regular sampling towards the highly populated classes is an {\it advantage\/} for the learning of
the structure parameters $\{\mathbf{\delta}_j\}$. After all, the point is exactly to learn these parameters from classes with substantial data and transfer them to the few-shot classes, where they cannot be learned accurately. Since the parameters are shared, both goals are accomplished if the learning procedure emphasizes the highly populated classes, as is the case for regular sampling. This implies that GIST training should include a mix of regular sampling (for shared structure parameters) and class-balanced sampling (for class specific parameters).

We propose to implement this with the hybrid training scheme of Figure~\ref{fig:structure}. In each iteration, two mini-batches $\mathbb{S}_c$, $\mathbb{S}_r$ are first sampled from the training set $\mathbb{S}$ by class-balanced sampling and random sampling, respectively. Two sets of class-specific parameters,
$\{\mathbf{w}_k,\bm{\upnu}_k\}$ are then learned, using the combination of (\ref{eq:softmaxce}),~(\ref{eq:amplifier}), and (\ref{eq:cosine}). The class-balanced mini-batch $\mathbb{S}_c$ is used with the resulting loss 
\begin{eqnarray}
    {\cal L}_c &=& \sum_{(\mathbf{x}_i, y_i)\in \mathbb{S}_c} \{-\max_j s(f_\phi(\mathbf{x}_i),\mathbf{w}_{{y_i}j}) \nonumber \\ 
    &&+\log \sum_{k}\exp[\max_j s(f_\phi(\mathbf{x}_i),\mathbf{w}_{kj})]\}, \nonumber \\
    \mathbf{w}_{kj} &=& g(\mathbf{w}_k, \mathbf{\delta}_j)
    \label{eq:loss_classbalance}
\end{eqnarray}
to learn the parameters $\mathbf{w}_k$.
The random mini-batch $\mathbb{S}_r$ is used with the loss
\begin{eqnarray}
    {\cal L}_r &=& \sum_{(\mathbf{x}_i, y_i)\in \mathbb{S}_r} \{-\max_j s(f_\phi(\mathbf{x}_i),\bm{\upnu}_{{y_i}j}) \nonumber \\ 
    && +\log \sum_{k}\exp[\max_j s(f_\phi(\mathbf{x}_i),\bm{\upnu}_{kj})]\}, \nonumber \\
    \bm{\upnu}_{kj} &=& g(\bm{\upnu}_k, \mathbf{\delta}_j)
    \label{eq:loss_random}
\end{eqnarray}
to learn the parameters $\bm{\upnu}_k$.
This results in the overall loss
\begin{equation}
    {\cal L} = {\cal L}_r + \lambda {\cal L}_c.
    \label{eq:loss_overall}
\end{equation}

The structure parameters $\mathbf{\delta}_j$ are common to the two losses. However, as shown in Figure~\ref{fig:structure}, during back-propagation only the gradient from ${\cal L}_r$ is used to update these parameters. This guarantees that the geometric structure is learned with random sampling. This structure is, however, propagated to the learning of the class specific parameters $\mathbf{w}_k$, which receive the gradient ${\cal L}_c$. In this way, the class specific parameters $\mathbf{w}_k$ are learned with class-balanced sampling, but this learning is informed by the structure parameters $\mathbf{\delta}_j$ learned with random sampling. This leads to parameter constellations $\mathbf{w}_{kj}$ that, while shared across classes, are centered at class-specific locations. 

Note that the displacements are forwarded together with $\mathbf{w}_k$ to calculate the class-balanced loss ${\cal L}_c$. This makes the two components $\{\mathbf{w}_j\}$ and $\{\delta_j\}$ of the classifier matching each other, although they are learned by different losses. The parameters $\bm{\upnu}_k$ are only used at training time, to guarantee that the geometric parameters $\mathbf{\delta}_j$ follow the natural geometry of the highly populated classes. They are discarded after training.

In GIST training, the class-specific weights $\mathbf{w}_k$ are trained with random sampling, while the structure parameters $\mathbf{\delta}_j$ are trained with class-balanced sampling. This forces the latter to predominantly represent the structure of the popular classes and is what enables the geometric structure transfer of Figure~\ref{fig:motivation}.

\begin{table*}
  \caption{Ablation of GistNet components, on the ImageNet-LT {\it validation set}. For many-shot $t >100$, for medium-shot $t \in (20,100]$, and for few-shot $t \leq 20$, where $t$ is the number of training samples.}
  \label{tab:ablation}
  \centering
  \begin{tabular}{lcccc}
    \toprule
    Method   & Overall & Many-Shot & Medium-Shot & Few-Shot \\
    \midrule
    Plain Model & 25.1 & 42.9 & 16.6 & 0.43 \\
    COS+CB & 37.6 & 49.4 & 34.8 & 14.7 \\
    COS+CS+CB & 39.5 & 52.6 & 36.3 & 14.5 \\
    COS+CS+GIST (GistNet) & {\bf 43.5} & {\bf 54.8} & {\bf 41.0} & {\bf 21.4} \\
    \midrule
    COS+GIST & 40.2 & 51.4 & 37.4 & 19.0 \\
    COS+CS+GIST ($\mathbf{w}_k$ and $\bm{\upnu}_k$ combined) & 40.9 & {\bf 58.2} & 34.6 & 14.8 \\
    COS+CS+GIST ($g$ rotation) & {\bf 43.6} & 55.1 & 40.8 & {\bf 21.7} \\
    COS+CS+GIST ($g$ MLP) & 43.4 & 54.2 & {\bf 41.1} & 21.5 \\
    \bottomrule
  \end{tabular}
\end{table*}

\begin{figure*}[t!]
\centering{
		\begin{tabular}{cccc} 
		\includegraphics[width=0.22\textwidth]{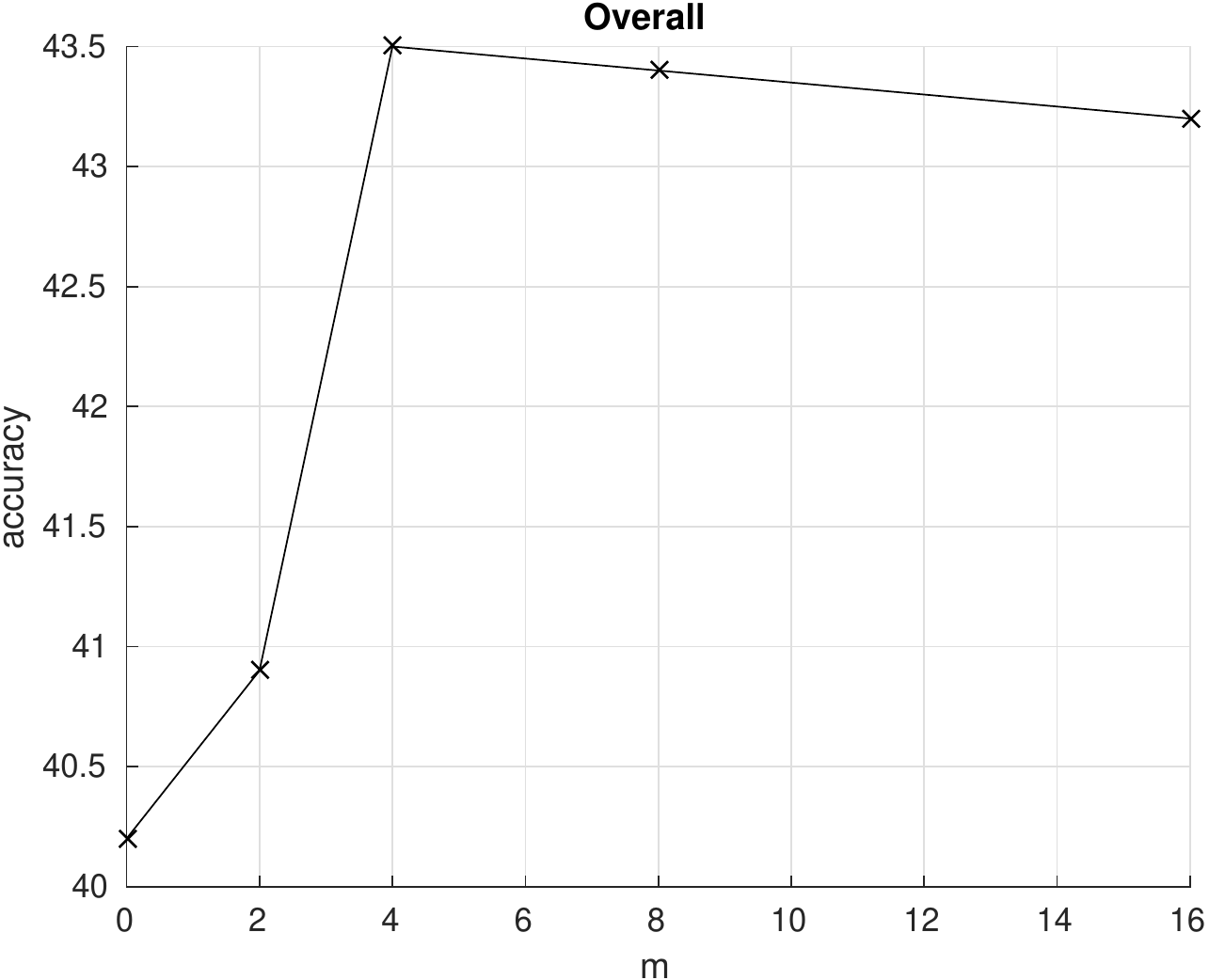} &
		\includegraphics[width=0.22\textwidth]{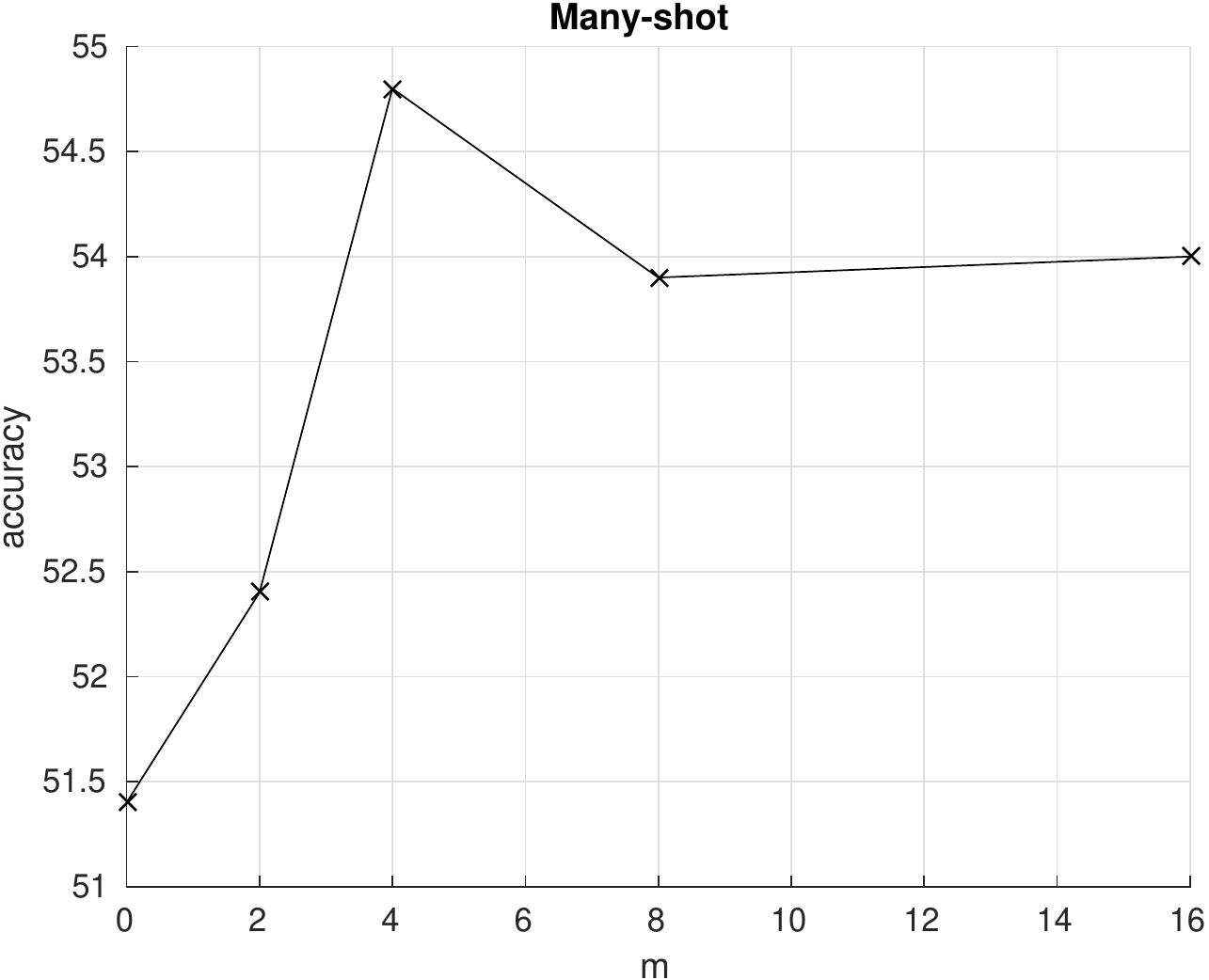} &
		\includegraphics[width=0.22\textwidth]{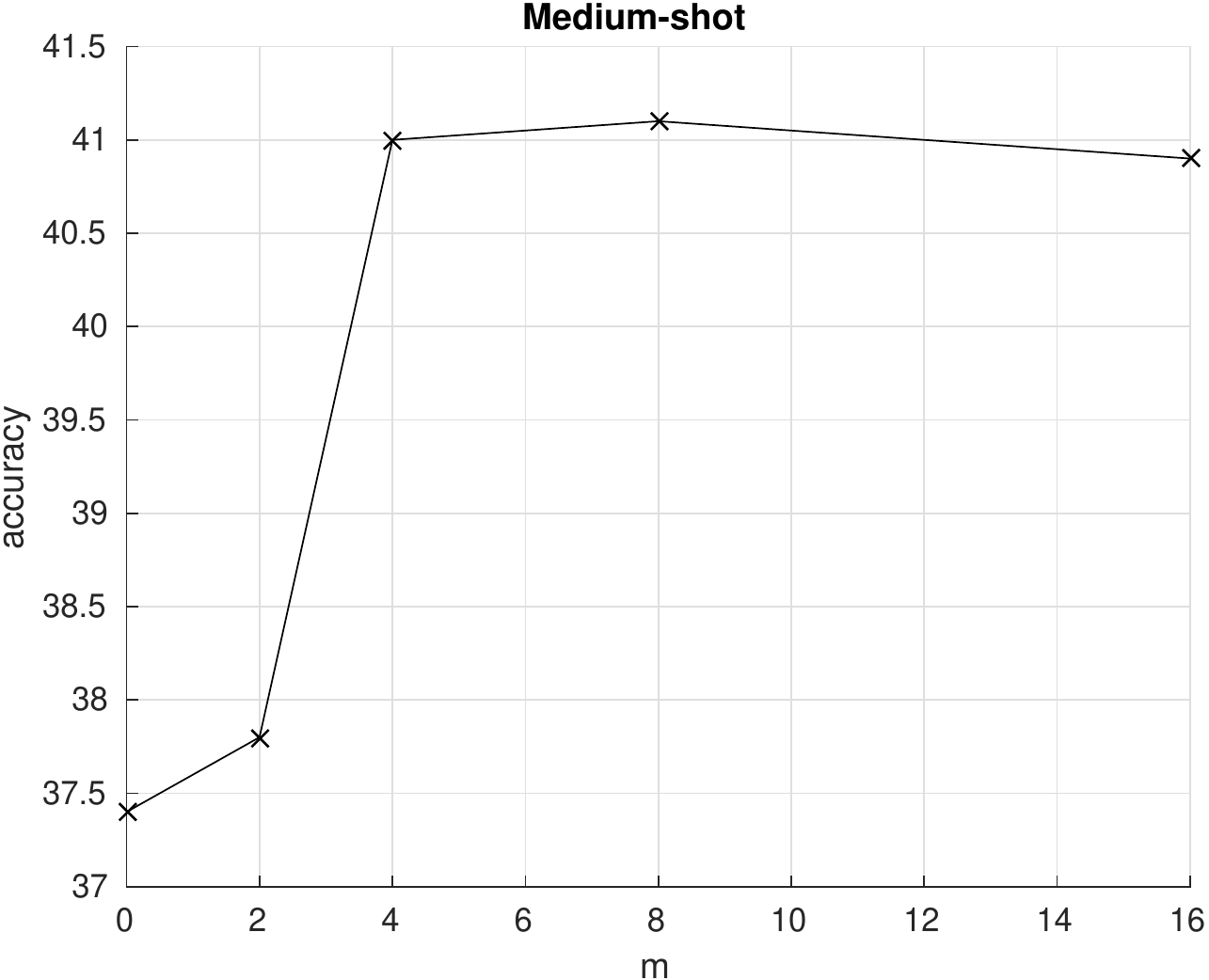} &
		\includegraphics[width=0.22\textwidth]{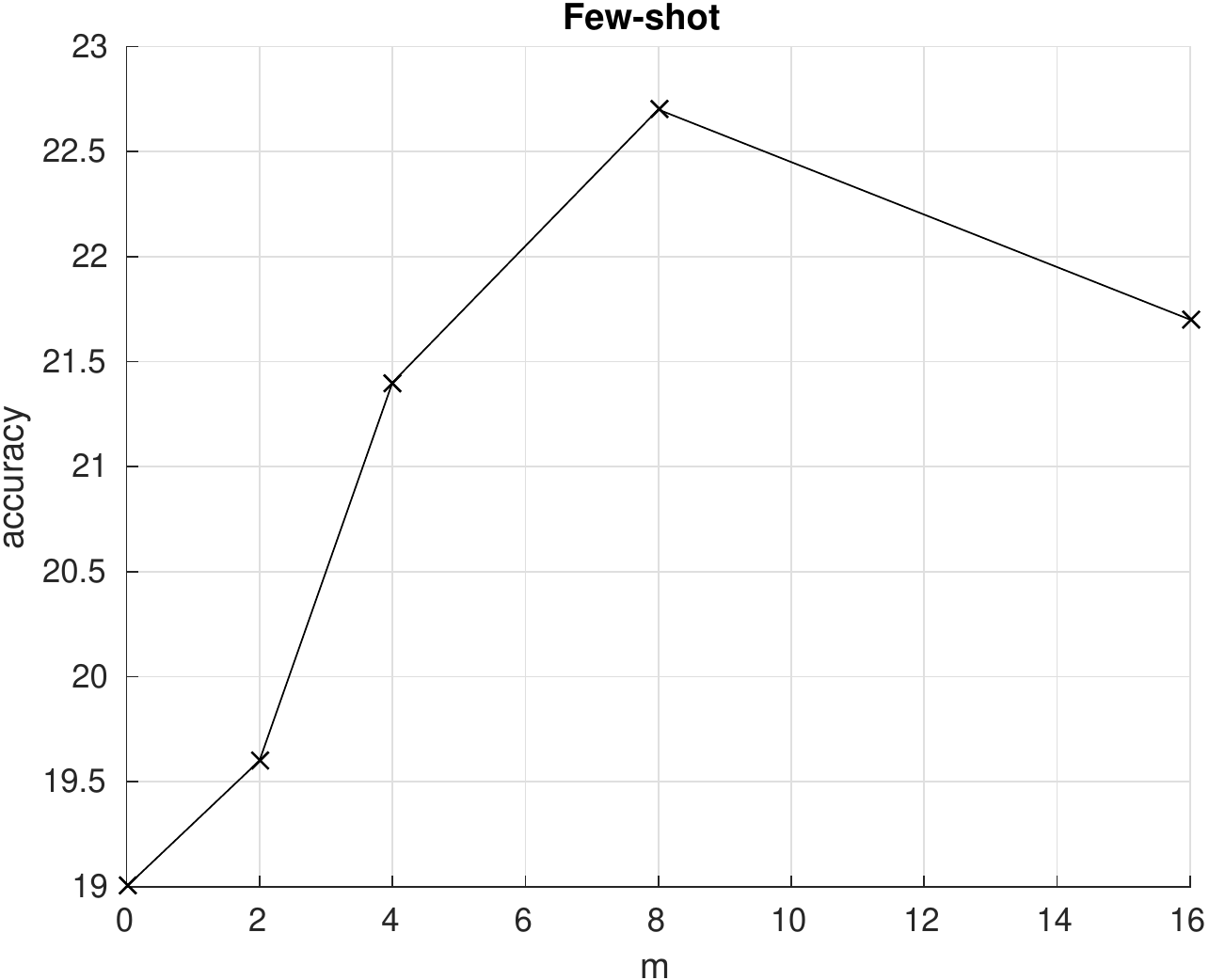} \\
		Overall & Many-shot & Medium-shot & Few-shot
		\end{tabular}}
	\caption{Results on different size of structure parameters in few-shot, medium-shot, many-shot classes, and overall accuracy, searched on validation set.}
	\label{fig:structure_parameters}
\end{figure*}
\section{Experiments}

In this section, we discuss an evaluation of the long-tailed recognition performance of the GistNet.

\subsection{Experimental set-up}
\noindent{\bf Datasets.}
We consider three long-tailed recognition datasets, 
ImageNet-LT~\cite{liu2019large}, Places-LT~\cite{liu2019large} and iNatrualist18~\cite{van2018inaturalist}. ImageNet-LT is a long-tailed version of ImageNet~\cite{imagenet_cvpr09} by sampling a subset following the Pareto distribution with power value $\alpha=6$. 
It contains $115.8$K images from $1000$ categories, with class cardinality ranging from $5$ to $1280$.
Places-LT is a long-tailed version of the Places dataset~\cite{zhou2014learning}. 
It contains $184.5$K images from $365$ categories with class cardinality ranging from $5$ to $4980$.
iNatrualist18 is a long-tailed dataset, which contains $437.5$K images from $8141$ categories with class cardinality ranging from $2$ to $1000$.

\noindent{\bf Baselines.}
Following~\cite{liu2019large}, we consider three metric-learning baselines, based on the lifted~\cite{oh2016deep}, focal~\cite{lin2017focal}, and range~\cite{zhang2017range} losses, 
and one state-of-the-art method, FSLwF~\cite{gidaris2018dynamic}, for learning without forgetting.
We also include state-of-the-art long-tailed recognition methods designed specifically for 
these two datasets:  OLTR~\cite{liu2019large}, Decoupling~\cite{kang2019decoupling}, and Distill~\cite{xiang2020learning}. The classifier of~(\ref{eq:softmax}) with standard random sampling is denoted as the {\it Plain Model} for comparison.

\noindent{\bf Training Details.}
ResNet-10 and ResNet-152~\cite{he2016deep} are used on ImageNet-LT and Places-LT respectively, and ResNet-50 is used on iNatrualist18.
Unless otherwise noted, we use four vectors $\delta_j$ of structure parameters, each with the dimension of $f_\theta(\mathbf{x})$. The class center $\mathbf{w}_k$ completes a constellation of five vectors. The number of structure parameters is ablated in Section~\ref{sec:ablate}. In all experiments, $\lambda=0.5$ is used in (\ref{eq:loss_overall}). 

The model is first pre-trained without structure parameters, with $60$ epochs of SGD, using momentum $0.9$, weight decay $0.0005$, and a learning rate of $0.1$ that decays by $10\%$ every $15$ epochs. After this, the full model is subject to GIST training with momentum $0.9$, weight decay $0.0005$ for $60$ epochs, and learning rate $0.1$ that decays by $10\%$ every $15$ epochs. In this case, each iteration uses class-balanced and random sampling mini-batches of size $128$, for an overall batch size of $256$. One epoch is defined when the random sampling iterates over the entire training data. Codes are attached in supplementary.

\subsection{Results}
Table~\ref{tab:imagenetlt} present results on ImageNet-LT and Places-LT. GistNet outperforms all other methods on the two datasets. A further comparison is performed by splitting classes into {\it many shot} (more than $100$ training samples), {\it medium shot} (between $20$ and $100$ training samples), and {\it few shot} (less that $20$ training samples). GistNet achieves the best performance on $5$ of the $6$ partitions and is competitive on the remaining one.

While on Places-LT the largest gains are for the few-shot classes, in ImageNet-LT they hold for the medium- and many-shot classes. 
This suggests that, in this dataset, the remaining methods overfit to the few-shot classes. The higher robustness of GistNet to this overfitting can be explained by the predominance of the many-shot classes in the training of the structure parameters $\mathbf{\delta}_j$. Results on iNaturalist18 dataset are shown in Table~\ref{tab:inat}, ours also outperforms all other methods.

\begin{figure*}[t!]
\centering{
		\begin{tabular}{ccc} 
		\includegraphics[width=0.31\textwidth]{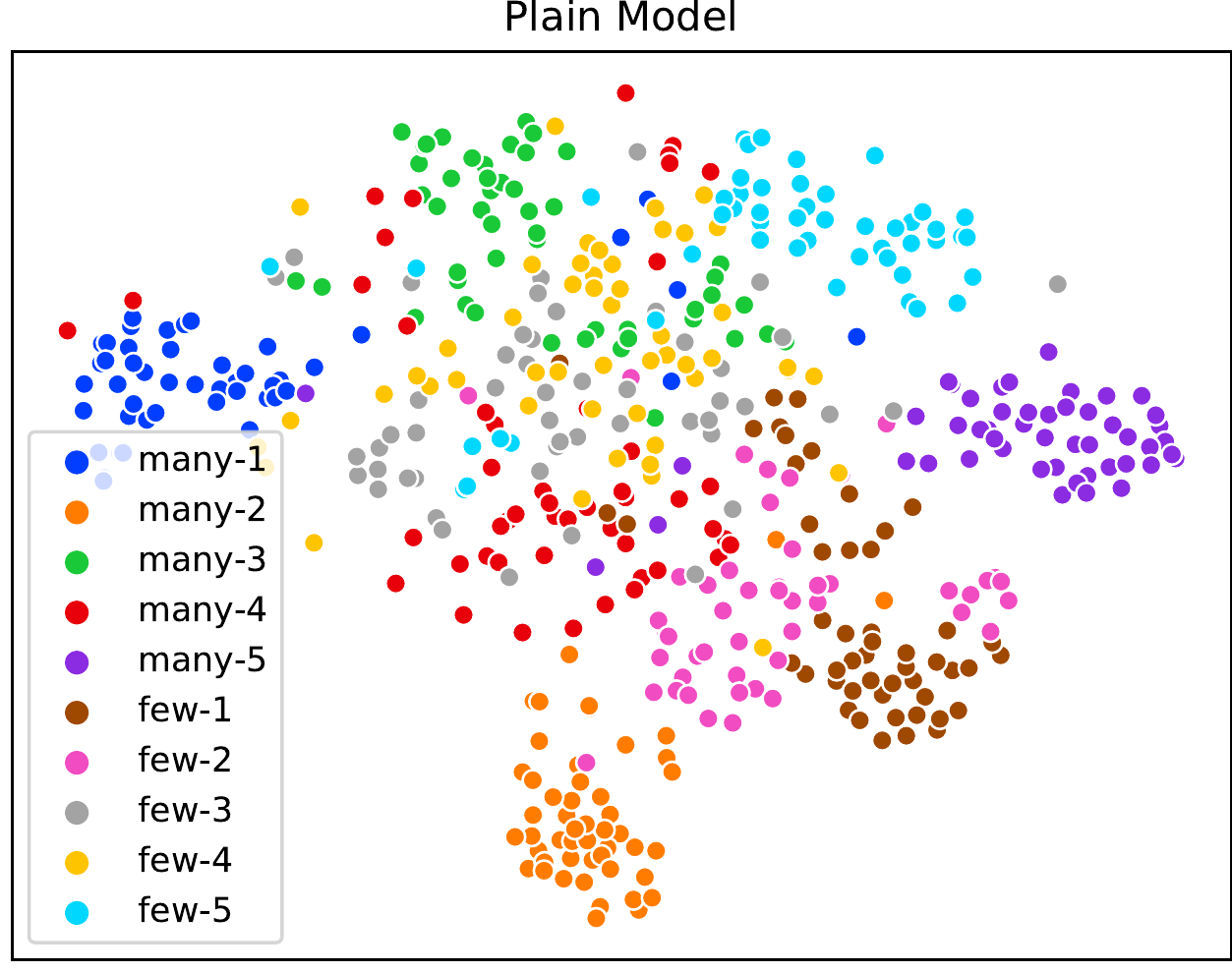} &
		\includegraphics[width=0.31\textwidth]{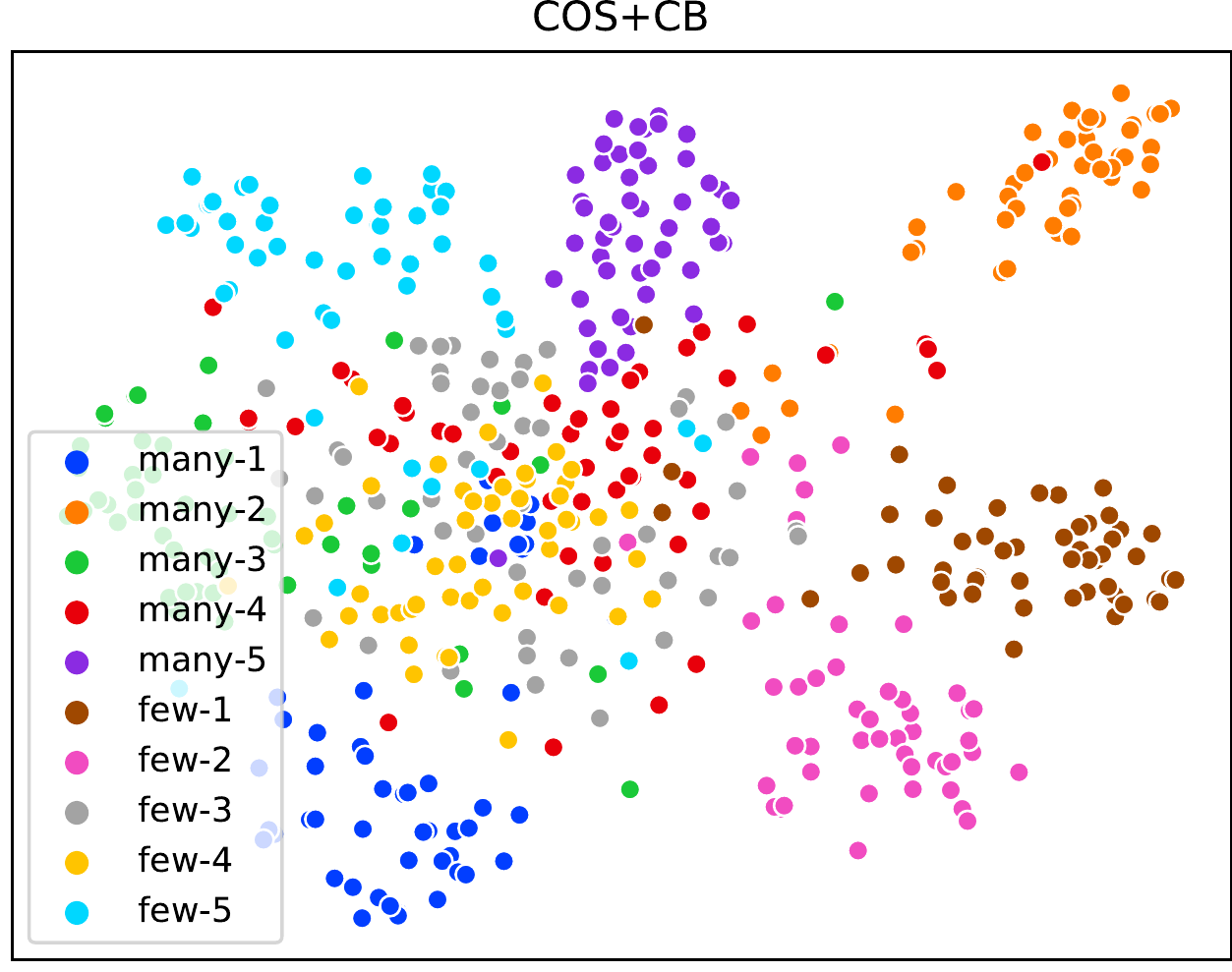} &
		\includegraphics[width=0.31\textwidth]{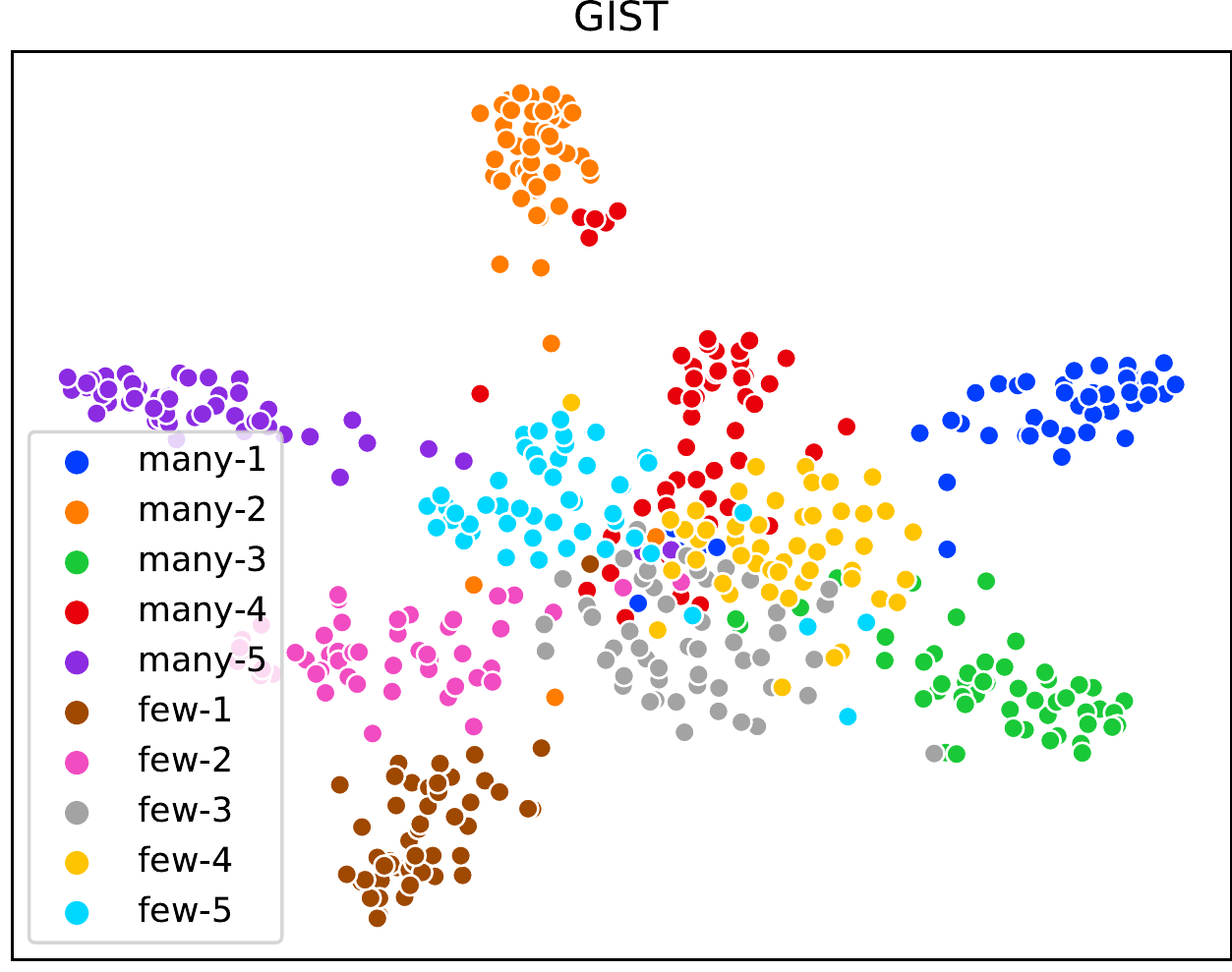} \\
		\footnotesize{Plain Model} & \footnotesize{COS+CB} & \footnotesize{GistNet}
		\end{tabular}}
	\caption{t-SNE visualizations of the embedding of {\it test set} images from $5$ randomly chosen many- and few-shot ImageNet-LT classes, for three models.}
	\label{fig:tsne}
\end{figure*}

\subsection{Ablation Study}
\label{sec:ablate}

In this section, we discuss the effectiveness of the various components of GistNet, the choice of constellation function $g$, the number of structure parameters, and the actual usage of constellations. All models are trained and evaluated on the training and validation set of ImageNet-LT, respectively, using a ResNet-10 backbone.

\noindent{\bf Component ablation.}
Starting from the plain model of~(\ref{eq:softmax}), we incrementally add the cosine classifier (COS) used in~(\ref{eq:cosine}), class-balanced sampling (CB), class structure parameters (CS), and GIST training (GIST). Table~\ref{tab:ablation} shows that the  combination of cosine classifier and class-balanced sampling (COS+CB) improves significantly on the plain model. The simple addition of the class structure parameters (COS+CS+CB) further improves the overall performance. 

However, there is no significant improvement for few-shot classes. This can be explained by the fact that, with class-balanced sampling, the three class types are equally predominant for the learning of the structure parameters. Hence, there is no transfer of geometric structure from many- to few-shot classes. This is confirmed by the fact that, when GIST training is added (COS+CS+GIST), performance improves significantly for the  few-shot classes. When compared to COS+CB, the GistNet model (COS+CS+GIST) has an overall gain of about $6$ points and better performance for all class types. Among these, the gains are particularly large (around $6.5$ points) for the few-shot classes.

The middle of the table investigates other possible configurations of the GistNet. Applying GIST training without class structure parameters (COS+GIST), i.e. using the combination of class balanced and random sampling only to learn the embedding $f_\phi(\mathbf{x})$, degrades performance for all class partitions. This shows the importance of enforcing a shared class structure among all classes. 

Another variant is to remove the additional class centers $\{\bm{\upnu}_k\}$ of Figure~\ref{fig:structure}, using the centers $\{\mathbf{w}_k\}$ for both losses, i.e. replacing $\bm{\upnu}_k$ with $\mathbf{w}_k$ in (\ref{eq:loss_random}). This variant, denoted COS+CS+GIST ($\mathbf{w}_k$ and $\bm{\upnu}_k$ combined), eliminates all the gains of GistNet for few-shot classes, while increasing the recognition accuracy for those in the many-shot partition. This is because the centers now receive gradient from the random sampling loss and are predominantly trained with many-shot data. The improved performance of GistNet over this variant shows that it is important to maintain the class-specificity of center training, while enforcing transfer of the geometric structure parameters, as done by GIST.

\noindent{\bf Different choices of $g$.} Beyond these variants, we have also considered different choices for the function $g$ that defines the parameter constellations of~(\ref{eq:amplifier}). In addition to the default addition function implemented by GistNet, we considered two possibilities. 

The first was a rotation. After the embedding and classifier parameters are normalized, we evaluate the distance between them on the $d$-dimensional unit sphere (where $d$ is the dimension of $f_\phi(\mathbf{x})$). The structure parameters are then $d$-dimensional rotation matrices, which encourage all classes to have the same structure on the unit sphere. This is implement the rotation matrix by a transformation of $d$-dimensional displacement vector
\begin{equation}
    \mathbf{R} = \mathbf{I}-\mathbf{u}\mathbf{u}^T-\mathbf{v}\mathbf{v}^T + [\mathbf{u}, \mathbf{v}]\mathbf{R}_{\theta}[\mathbf{u}, \mathbf{v}]^T,
    \label{eq:R}
\end{equation}
where $\mathbf{u}$ is a unit vector, $\mathbf{v}$ is the normalized vector of a displacement vector $\delta_j$, and $\mathbf{R}_{\theta}$ is the 2D rotation matrix between $\mathbf{u}$ and $\delta$. Given a structure parameter vector $\delta_j$, the parameter constellations are implemented as
\begin{equation}
    \mathbf{w}_{kj} = g(\mathbf{w}_k, \delta_j) = \mathbf{R} \mathbf{w}_k
\end{equation}
Details are discussed in supplementary. 

The second was a learned function $g$, implemented by a two-layer MLP, and learned end-to-end. 

Table~\ref{tab:ablation} shows that the different implementations of $g$ have little impact on the recognition performance. This suggests that the addition of global geometric constraints is much more important than the specific implementation details of these constraints.

\noindent{\bf Number of structure parameters.}
We next investigated the influence of the number $m$ of structure parameters $\{\delta_j\}_{j=1}^m$. As shown in Figure~\ref{fig:structure_parameters}, none of the alternatives tried ($m \in \{2, 8, 16\}$) outperformed the four parameters used in GistNet. For overall, many-, and medium-shot classes performance increases until $m=4$ and then saturates. For few-shot classes, there was a one-point gain in using $m=8$. This shows that this partition is the one that most benefits from geometry transfer. 

Overall, these results confirm that while geometric transfer can produce significant gains, the GistNet architecture is quite robust to variations of detail.

\noindent{\bf Actual usage of constellations.}
Cross-entropy minimization encourages the use of more $\delta_j$, since the coverage of the class distributions is better. It would be a waste not to use them all. In the test set of ImageNet-LT, the actual usage was $\{25\%, 23\%, 18\%, 17\%, 17\%\}$. $792$ of $1000$ classes chose each $\delta_j$ for at least $10\%$ of test samples. This results further support that the model does not collapse to traditional classifier by fitting to only one constellation and ignoring others. 

\subsection{Visualization}
Figure~\ref{fig:tsne} shows a t-SNE~\cite{maaten2008visualizing} visualization of the embeddings learned by the Plain Model, the COS+CB baseline, and GistNet. For clarity, we randomly choose five classes from the many- and few-shot splits in ImageNet-LT. The figure shows the t-SNE projection of features of {\it test} samples from those classes. Compared to the two other models, GistNet produces classes that are better separated and have more consistent geometry. This is especially true for few-shot classes.

\section{Conclusion}

This work addressed the long-tailed recognition problem. A new architecture, GistNet, and training scheme, GIST, were proposed to enable transfer of geometric structure from highly populated to low-populated classes. This leverages the tendency of SGD training to overfit to the populated classes, rather than simply fighting this tendency. 

GistNet was shown to achieve state-of-the-art performance on two popular long-tailed datasets. Ablation studies have shown that, while geometric transfer enables significant recognition gains, the architecture is quite robust to variations of detail. This suggests that the addition of global geometric constraints to long-tailed recognition is more important than the specific implementation of these constraints.


{\small
\bibliographystyle{ieee_fullname}
\bibliography{Gist}
}

\end{document}